\documentclass{article}
\usepackage[preprint]{nips}

\usepackage{graphicx}
\usepackage[labelfont=bf]{caption}
\usepackage[superscript,nomove]{cite}
\usepackage{color}
\usepackage{amsmath}
\usepackage{amsfonts}
\usepackage{amssymb}
\usepackage{subcaption}
\usepackage{todonotes}
\usepackage{enumitem}
\usepackage{gensymb}
\usepackage{textcomp}
\usepackage{hyperref}

\begin{document}

\title{Sepsis Prediction with \\ Temporal Convolutional Networks}

\author{
Xing Wang\\
Splunk Inc. \\
{\tt\small xingw@splunk.com}
\And 
Yuntian He\\
Georgia Institute of Technology \\
{\tt\small yhe353@gatech.edu}
}




\maketitle


\section*{Abstract}
We design and implement a temporal convolutional network model to predict sepsis onset. Our model is trained on data extracted from MIMIC III database, based on a retrospective analysis of patients admitted to intensive care unit who did not fall under the definition of sepsis at the time of admission. Benchmarked with several machine learning models, our model is superior on this binary classification task, demonstrates the prediction power of convolutional networks for temporal patterns, also shows the significant impact of having longer look back time on sepsis prediction.


\section{Introduction}
Sepsis is one of the leading causes of mortality in worldwide. The incidence and mortality rates failed to decrease over the last few decades. Early recognition and early aggressive treatment may improve the outcomes in sepsis. Early and accurate onset predictions could allow more aggressive and focused treatment, which could reduce the incidence of multi-organ dysfunction and significantly decreased the in-hospital mortality rate\cite{kim2019sepsis}.
Due to disease complexity in clinical context, early recognition can be difficult\cite{ediriweera2016mapping, talisa2018arguing}. Existing detection methods cannot provide high performance and real-time laboratory test results, which would lead the delay of treatment. With the development of machine learning as a tool to analyze large amount of data, prediction models can be developed based on machine learning to diagnose sepsis ahead of time, which may assist onset clinical to decrease the mortality.

Recurrent neural network (RNN) and its variants of sequence to sequence (Seq2Seq) framework \cite{sutskever2014sequence} have achieved great success in many sequential modeling tasks, such as machine translation \cite{cho2014learning}, speech recognition \cite{sak2014long}, natural language processing \cite{bengio2003neural}, and extensions to autoregressive time series forecasting \cite{salinas2019deepar, rangapuram2018deep} in recent years. 
And there have been work on applying RNN on sepsis prediction \cite{scherpf2019predicting}.
However, RNNs can suffer from several major challenges. Due to its inherent temporal nature (i.e., the hidden state is propagated through time), the training cannot be parallelized. Moreover, trained with backpropagation through time (BPTT) \cite{werbos1990backpropagation}, RNNs severely suffer from the problem of gradient vanishing thus often cannot capture long time dependency \cite{pascanu2013difficulty}. More elaborate architectures of RNNs use gating mechanisms to alleviate the gradient vanishing problem, 
the long short-term memory (LSTM) \cite{hochreiter1997lstm} and its simplified variant, the gated recurrent unit (GRU) \cite{chung2014gru}, are the two canonical architectures commonly used in practice.
On the other hand, convolutional neural networks (CNNs) \cite{lecun1989backpropagation} can be easily parallelized, and recent advances effectively eliminate the vanishing gradient issue and hence help building very deep CNNs. These works include the residual network (ResNet) \cite{he2016deep} and its variants such as highway network \cite{srivastava2015training}, DenseNet \cite{huang2017densely}, etc.
In the area of sequential modeling, 1D convolutional networks offered an alternative to RNNs for decades \cite{waibel1989phoneme}. 
In recent years, Oord et al.\cite{oord2016wavenet} proposed WaveNet, a dilated causal convolutional network as an autoregressive generative model. Ever since, multiple research efforts have shown that with a few modifications, certain convolutional architectures achieve state-of-the-art performance in 
various fields \cite{oord2016wavenet, binkowski2018autoregressive, chen2020probabilistic}.
In particular, Bai et al.\cite{bai2018empirical} abandoned the gating mechnism in WaveNet and proposed temporal convolutional network (TCN). The authors benchmarked TCN with LSTM and GRU on several sequence modeling problems, and demonstrated that TCN exhibits substantially longer memory and achieves better performance.
In this paper, we build a WaveNet-style TCN for learning temporal information from historical data, with the application in sepsis prediction.

The rest of this report is organized as follows. Section \ref{sec:method} formulate the problem and introduce the collection and preprocessing of the data. Section \ref{sec:model} illustrates the building blocks as well as the implementation and learning details of the TCN model. We benchmark our TCN with several other models, the evaluation results are shown in Section \ref{sec:res}. Finally, we conclude our findings and discuss future work directions in Section \ref{sec:conc}.

\section{Problem Formulation and Data} \label{sec:method}
We aim at making predictions of sepsis prediction onset. 
As our goal is to distinguish patients who obtained sepsis at any point in time during their stay in the intensive care unit from those who did not, two classes were defined: the sepsis-class and non-sepsis-class. The point in time from where sepsis is to be predicted is defined by the difference of sepsis onset and the prediction time.

    
    \subsection{Data Collection and Inclusion Criteria}
    The MIMIC III database was recorded between 2001 and 2012, in the Beth Israel Deaconess Medical Center in Boston, Massachusetts. We use the most recent Version (v1.4) for this work. The database contains 58976 admissions of 46520 patients. 
    
    After getting all the data we needed in this paper, we apply the tool from MIT-LCP to get the pivoted SOFA table, which extracts the sequential organ failure assessment for the patients in the ICU. SOFA\cite{singer2016third} score is used to track a person's status during the stay in an intensive care unit (ICU) to determine the extent of a person's organ function or rate of failure. The score is based on six different scores, one each for the respiratory, cardiovascular, hepatic, coagulation, renal and neurological systems. The score is calculated every hour during the patient's ICU stay. Meanwhile, the tool\cite{johnson2018comparative} was applied to get the suspicious infection time and sepsis cohort. The parameters that the query used are listed in Table 1.
    
    \subsection{Data pre-processing}
    From the result of the suspicious infection time, sepsis cohort and pivoted SOFA table, we apply our sepsis-label criteria. We observed the SOFA score from 2 days before the suspicious infection time to 1 day after the suspicious infection time. If the score went up by 2 in the time window, the patients will be labeled as sepsis onset. Meanwhile, the time window need to be between ICU in time and ICU out time.
    
    \begin{figure}[htb]
    \centering
    \centering
    \includegraphics[width=0.6\textwidth]{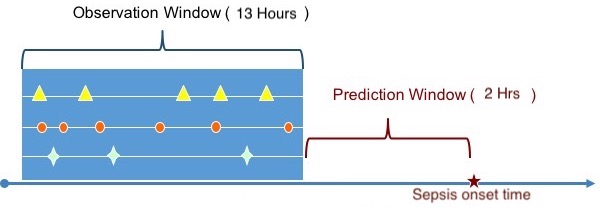}
    \caption{Observation Window/Prediction Window}
    \label{fig: Window}
    \end{figure}
    
    The look back is the sequence of values that is used to predict if there will occur a sepsis onset or not, so if a specific look back is classified as belonging to the sepsis-class or non-sepsis-class. The look back time here we applied is 13 hours, and prediction window is 2 hours. The length of look back time has a significant impact on the performance of the classifier. Jozwiak et al.\cite{jozwiak2016implementing} indicates the earlier the treatment is achieved, the lower is the mortality. It has been shown that the compliance with early resuscitation treatments, within the first 3 hours, was associated with a lower probability of being eligible for later resuscitation and maintenance bundle element. 
    
    \begin{table}[bth!]
        \centering
        \footnotesize
        \begin{tabular}{|c|c|}
            \hline
            Parameters & ITEMIDs \\
            \hline
            patient age &  $\ge$ 18\\
            \hline
            systolic blood presure & 220050, 225309, 220179, 51, 455, 6701\\
            \hline
            diastolic blood pressure & 220051, 225309, 225310, 8386, 8441, 8555\\
            \hline
            blood oxygen saturation($SO_{2}$) & 220227, 220277, 834, 646\\
            \hline
            temperature & 223762, 223761, 676, 678\\
            \hline
            heart rate & 220045, 211\\
            \hline
            respiratory rate & 220210, 224422, 224689, 224690, 618, 651, 615, 614\\
            \hline
            $CO_{2}$ partial pressure($PaCO_{2}$) & 220235, 778\\
            \hline
        \end{tabular}
        \caption{\label{tab:extract parameter}Extracted parameters from the MIMIC III database. The ITEMID represents the identification number of a measurement in the MIMIC III database.}
    \end{table}
    
\begin{figure}[htb]
    \centering
    \includegraphics[width=.7\textwidth]{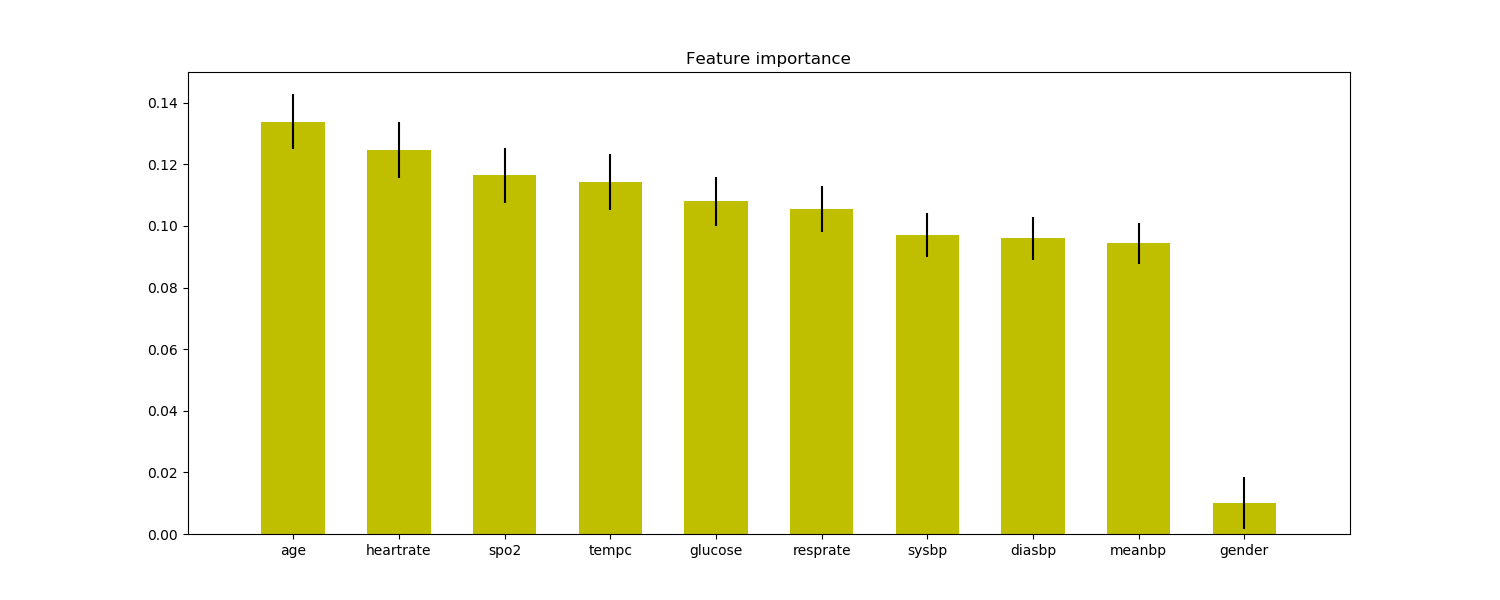}
    \vspace{-.1in}
    \caption{Feature importance plot of tree model.}
    \label{fig:feature_importance}
\end{figure}

As a tentative illustration, Figure \ref{fig:feature_importance} shows the most important 10 features for predicting the sepsis onset, according to the decision tree model we trained for benchmarking.

\section{Approach and Implementation} \label{sec:model}
\subsection{TCN Architecture}
As we mentioned above, we will use temporal CNN to replace the temporal module learns from historical data, where RNNs are usually applied. 
CNNs have shown success in time series applications, in which the 1D convolution is simply an operation of sliding dot products between the input vector and the kernel vector. We will make several modifications to traditional 1D convolutions according to recent advances. 
The detailed building blocks of our temporal CNN components are illustrated in the following  subsections.
\subsubsection{Causal Convolutions} \label{sec:causal}
\begin{figure}[htb]
    \centering
    \begin{subfigure}[b]{0.457\textwidth}
        \centering
        \includegraphics[width=0.6\textwidth]{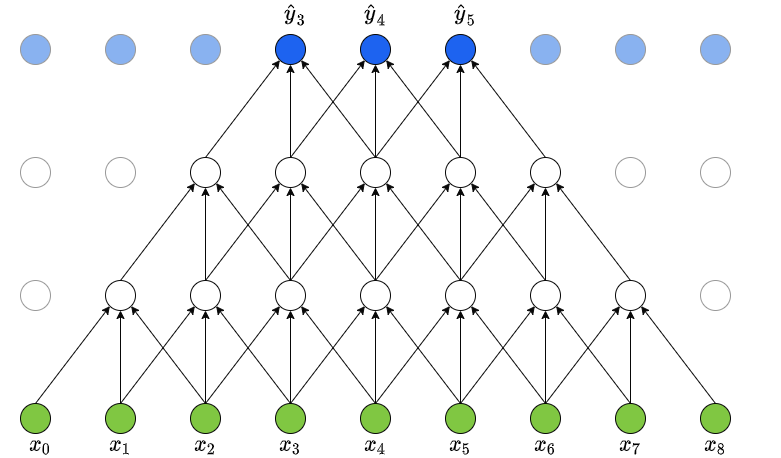}
        \caption{standard (non-causal)}
        \label{fig:causal}
    \end{subfigure}
    ~
    \begin{subfigure}[b]{0.475\textwidth}
        \centering
        \includegraphics[width=0.6\textwidth]{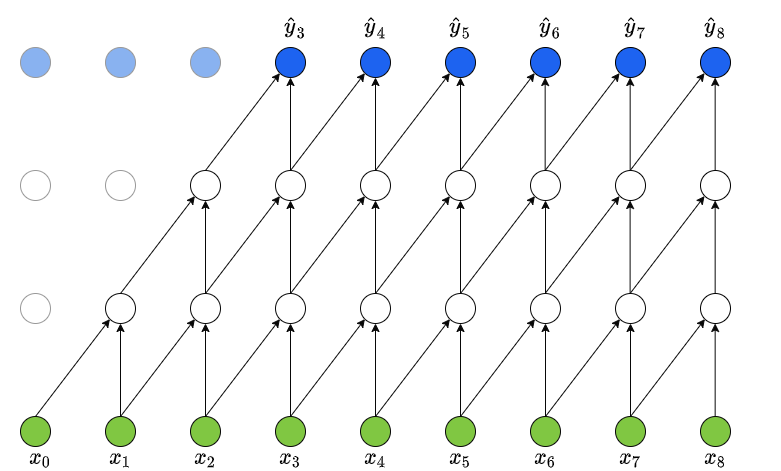}
        \caption{causal}
    \end{subfigure}
    \caption{Visualization of a stack of 1D convolutional layers, non-causal v.s. causal.}
    \label{fig:non_vs_causal}
\end{figure}

In a traditional 1D convolutional layer, the filters are slided across the input series. As a result, the output is related to the connection structure between the inputs before and after it. As shown in Figure \ref{fig:non_vs_causal}(a), by applying a filter of width 2 without padding, the predicted outputs $\hat{x}_1, \cdots, \hat{x}_T$ are generated using the input series $x_1, \cdots, x_T$. The most severe problem within this structure is that we use the future to predict the past, e.g., we have used $x_2$ to generate $\hat{x}_1$, which is not appropriate in time series analysis. To avoid the issue, causal convolutions are used, in which the output $x_t$ is convoluted only with input data which are earlier and up to time $t$ from the previous layer. We achieve this by explicitly zero padding of length $(kernel\_size - 1)$ at the beginning of input series, as a result, we actually have shifted the outputs for a number of time steps. In this way, the prediction at time $t$ is only allowed to connect to historical information, i.e., in a causal structure, 
thus we have prohibited the future affecting the past and avoided information leakage. 
The resulted causal convolutions is visualized in Figure \ref{fig:non_vs_causal}(b).

\subsubsection{Dilated Convolutions} \label{sec:dilated}
\begin{figure}[tbh]
    \centering
    \begin{subfigure}[b]{0.45\textwidth}
        \centering
        \includegraphics[width=\textwidth]{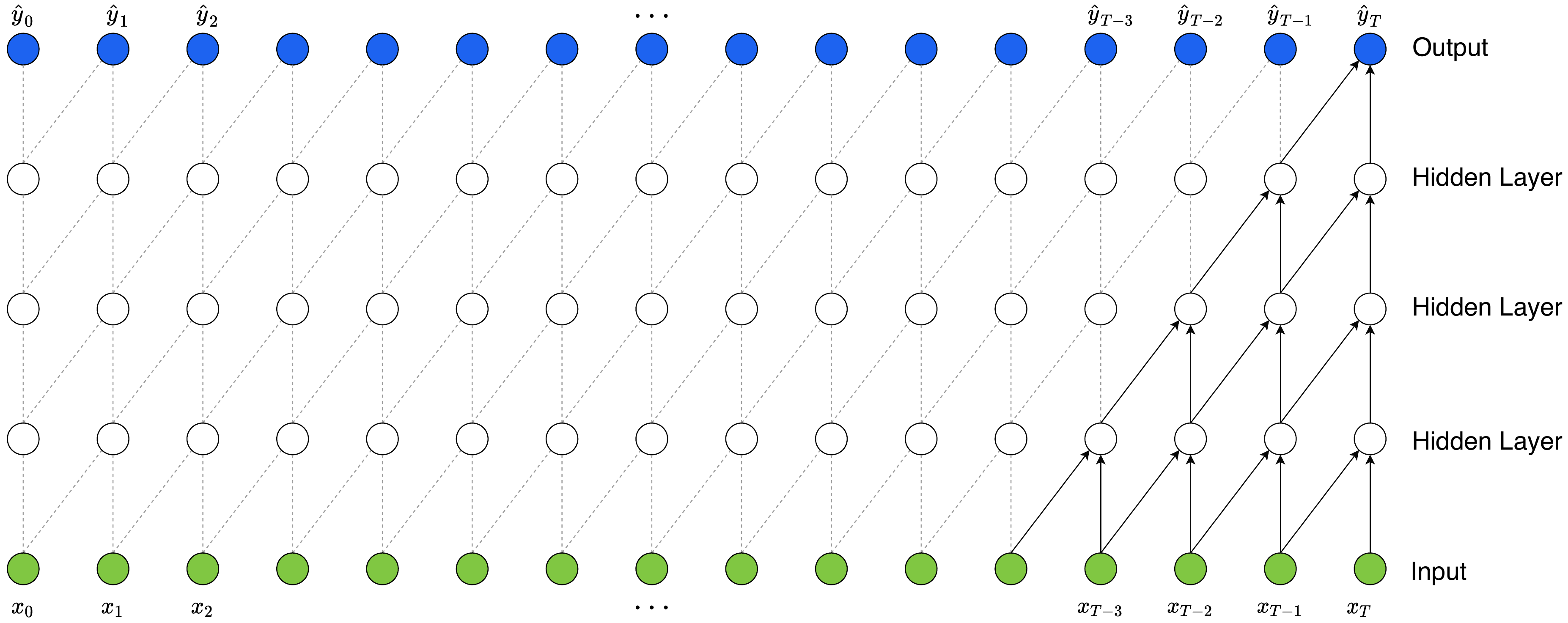}
        \caption{Non-dilated}
        \label{fig:nondilated}
    \end{subfigure}
    \hfill
    \begin{subfigure}[b]{0.45\textwidth}
        \centering
        \includegraphics[width=\textwidth]{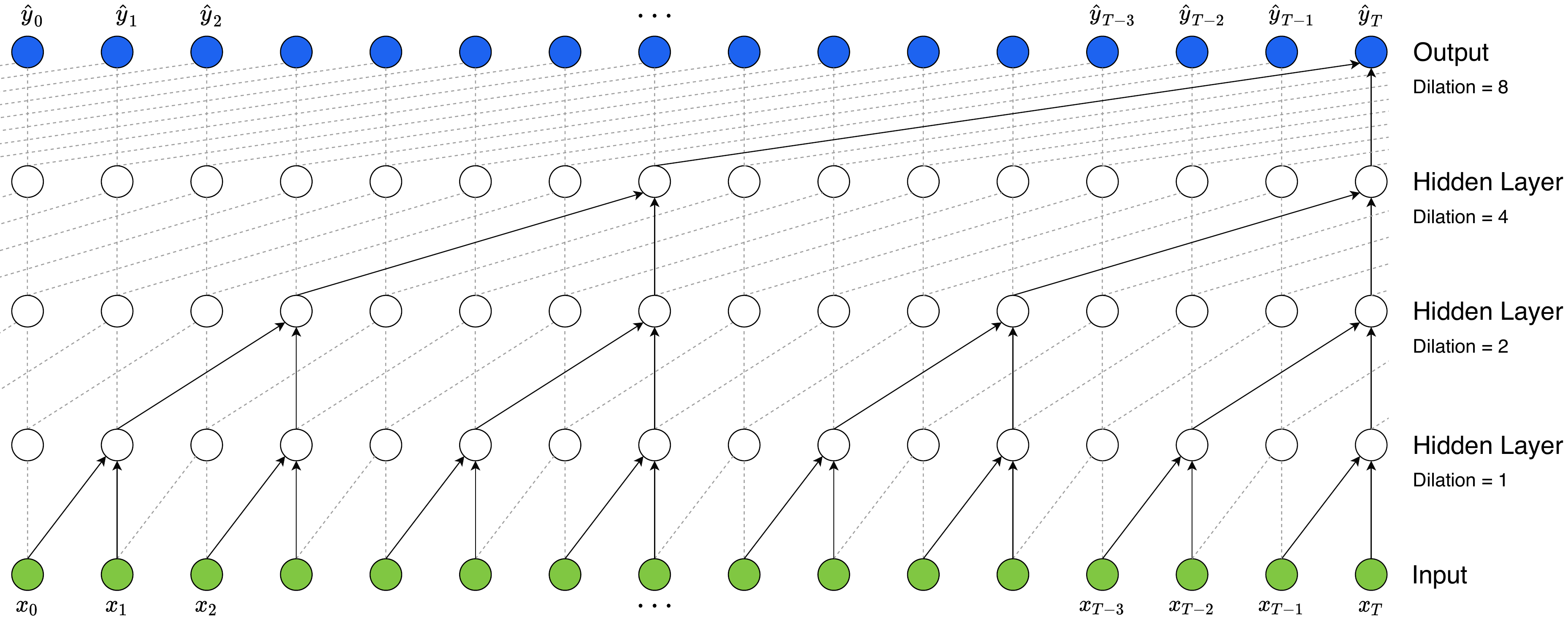}
        \caption{Dilated}
        \label{fig:dilated}
    \end{subfigure}
    \caption{Visualization of a stack of causal convolutional layers, non-dilated v.s. dilated.}
    \label{fig:non_vs_dilated}
\end{figure}

Time series often exhibits long-term autoregressive dependencies. With neural network models hence, we require for the receptive field of the output neuron to be large. That is, the output neuron should be connected with the neurons that receive the input data from many time steps in the past.  
A major disadvantage of the aforementioned basic causal convolution is that in order to have large receptive field,  either very large sized filters are required, or those need to be stacked in many layers. With the former, the merit of CNN architecture is  lost, and with the latter, the model can become computationally intractable.
Following Oord et al.\cite{oord2016wavenet}, we will adopt the dilated convolutions in our model instead, which is defined as $$F(s) = (\mathbf{x}\ast_d f)(s) = \sum_{i=0}^{k-1}f(i)\cdot \mathbf{x}_{s-d\times i}, $$ where $x\in \mathbb{R}^T$ is a 1-D series input, and $f:\{ 0, \cdots, k-1\} \to \mathbb{N}$ is a filter of size $k$, $d$ is called the dilation rate, and $(s-d\times i)$ accounts for the direction of the past. In a dilated convolutional layer, filters are not convoluted with the inputs in a simple sequential manner, but instead skipping a fixed number ($d$) of inputs in between. By increasing the dilation rate multiplicatively as the layer depth (e.g., a common choice is  $d=2^j$ at depth $j$), we increase the receptive field 
exponentially, i.e., there are $2^{l-1}k$ input in the first layer that can affect the output in the $l$-th hidden layer.  Figure \ref{fig:non_vs_dilated} compares non-dilated and dilated causal convolutional layers.

\subsubsection{Residual Connections} \label{sec:resnet}
In traditional neural networks, each layer feeds into the next. In a network with residual blocks, by utilizing skip connections, a layer may also short-cut to jump over several others. The use of residual network (ResNet) \cite{he2016deep} has been proven to be very successful and become the standard way of building deep CNNs. 
The core idea of ResNet is the usage of shortcut connection which skips one or more layers and directly connects to later layers (which is the so-called identity mapping), in addition to the standard layer stacking connection $\mathcal{F}$. 

\begin{figure}[htb]
    \centering
    \begin{subfigure}[b]{0.375\textwidth}
        \centering
        \includegraphics[scale=.5]{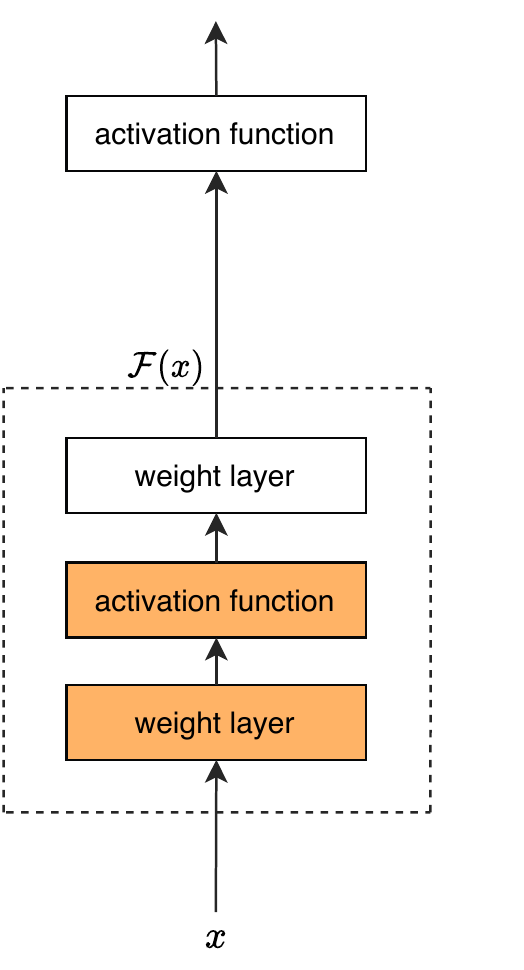}
        \caption{A standard block}
    \end{subfigure}
    ~
    \begin{subfigure}[b]{0.375\textwidth}
        \centering
        \includegraphics[scale=.5]{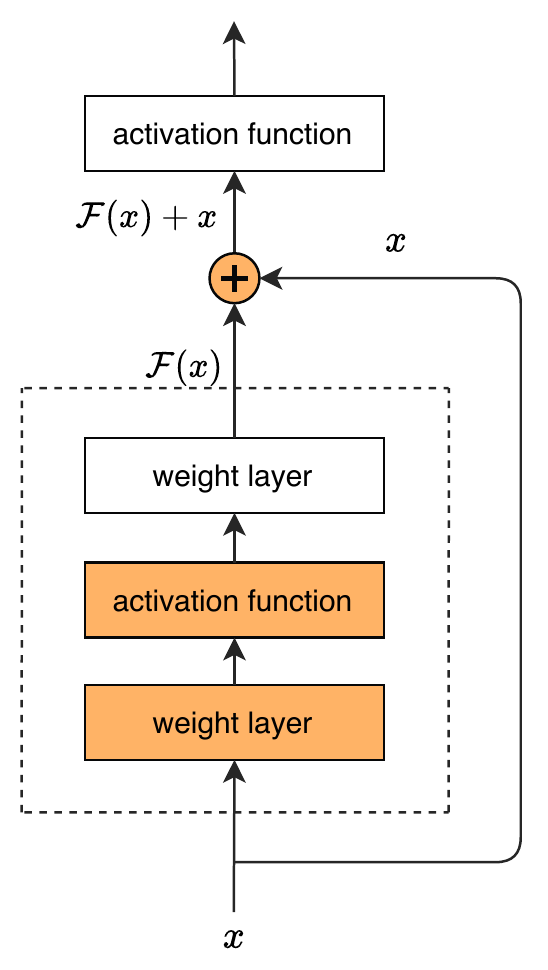}
        \caption{A residual block}
    \end{subfigure}
    \caption{Comparison between a regular block and a residual block. In the latter, the convolution is short-circuited.}
    \label{fig:resblock}
\end{figure}

Figure \ref{fig:resblock} illustrates a residual block, which is the basic unit in ResNet. A residual block consists of the abovementioned two branches, and its output is then $g(\mathcal{F}(x)+x)$, where $x$ denotes the input to the residual block, and $g$ is the activation function.
By reusing activation from a previous layer until the adjacent layer learns its weights, CNNs can effectively avoid the problem of vanishing gradients. 

\subsection{Implementation Details and Benchmark Models}
\begin{figure}[htb]
    \centering
    \includegraphics[width=0.8\textwidth]{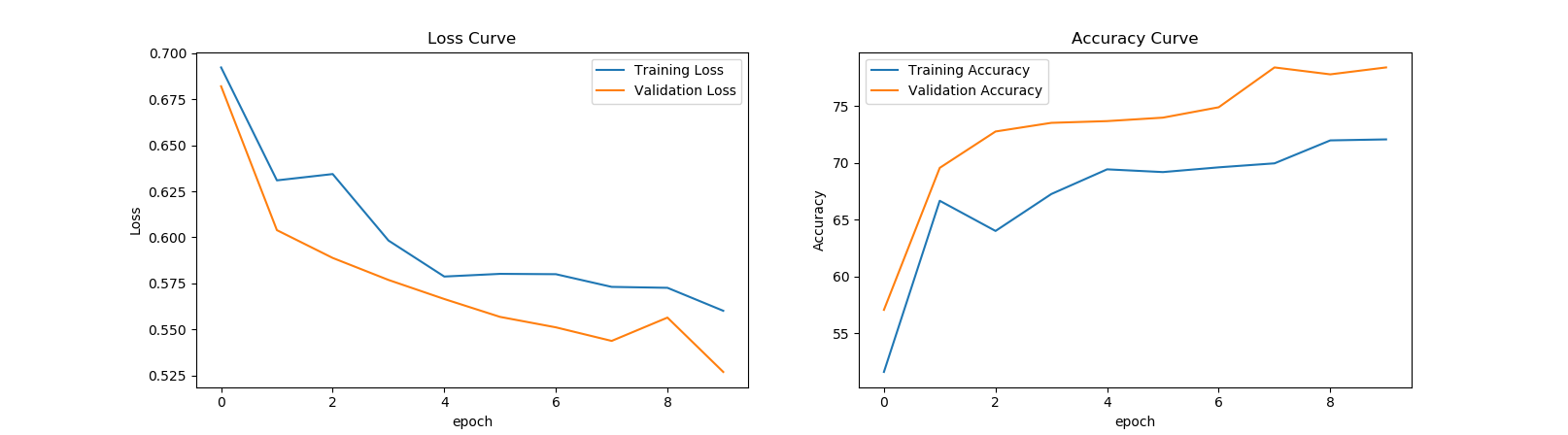}
    \caption{Learning curves of TCN.}
    \label{fig:lc}
\end{figure}

Our TCN model was implemented in PyTorch \cite{paszke2017automatic}. 
We build a 9-layer dilated causal CNN was built as the component that focuses on capturing the time-dependent features. Each layer contains 16 filters, and each filter has a width of 2. Every two consecutive convolutional layers form a residual block after which the previous inputs are added to the flow. The dilation rate increases exponentially along every stacked residual blocks, i.e., to be $1, 2, 4, 8, \cdots, 128$.
Within each TCN block, we use dropout (with probability 0.2) for the convolutional layers in order to limit
the influence that earlier data have on learning \cite{srivastava2014dropout}. It is then followed by a batch normalization layer \cite{ioffe2015batch}. Both dropout and batch normalization provide a regularization effect that avoids overfitting. The most widely used activation function, the rectified linear unit (ReLU) \cite{nair2010rectified} is used after each layer except for the last one. The convolutional output of the TCN module is a feature map which is set to be of size 4, it is then sent to a fully-connected classifier with 2 outputs, as the model is optimized on binary cross-entropy cost function, represents a standard approach for dichotomous classification tasks.
Note that the TCN module can be replaced by any other architecture that learns temporal patterns, for example, RNN-type network. As a benchmarking deep learning model, we built an LSTM model which consists of 2 hidden layers with 4 neurons each plays the role of the TCN module, while the rest of the model is the same as we mentioned above. 
The deep learning models were trained using stochastic gradient descent (SGD), with batch size of 64. In particular, the Adam optimizer \cite{kingma2014adam}  with initial learning rate of 0.005 was used for both LSTM and TCN. We trained the TCN and LSTM models for 10 epochs. Figure \ref{fig:lc} shows the changes of training and validation loss and accuracy over the number of epochs, we can see that as learning proceeds, the model performance keeps improving. 

In addition to LSTM, we also benchmark with several other machine learning models. 
Random forest \cite{breiman2001random} and AdaBoost \cite{freund1997decision} are ensemble models that deploy enhanced bagging and boosting, respectively. We pick these two models since both have shown powerful predicting ability. Scikit-learn \cite{pedregosa2011scikit} was used to build these machine learning models, and both ensemble models apply decision trees as the base learners. 
We split the data into training and testing set by a ratio of 80\% and 20\%, and the optimal hyperparameters of our traditional machine learning models were obtained with K-fold cross validation based on grid search. For the two deep learning models, we further split the training data into training and validation sets, and tuned the hyperpameters such as learning rate as well as hidden size, etc. according to the performance on validation data. 

\section{Experiment Results} \label{sec:res}

\begin{figure}[htb]
    \centering
    \begin{subfigure}[b]{0.475\textwidth}
        \centering
        \includegraphics[width=\textwidth]{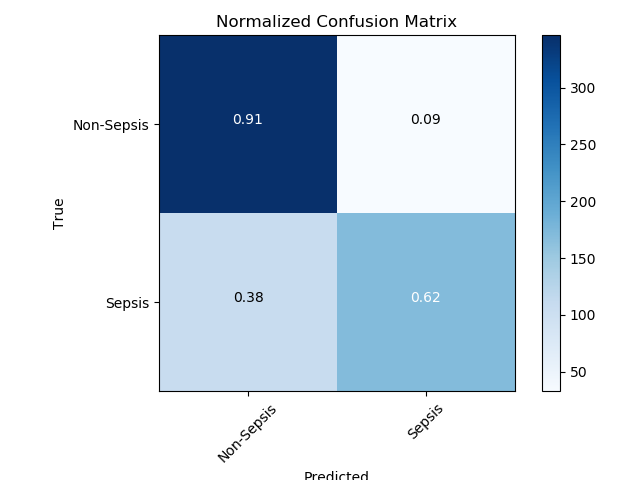}
    \caption{Confusion matrix}
    \end{subfigure}
    ~
    \begin{subfigure}[b]{0.475\textwidth}
        \centering
        \includegraphics[width=\textwidth]{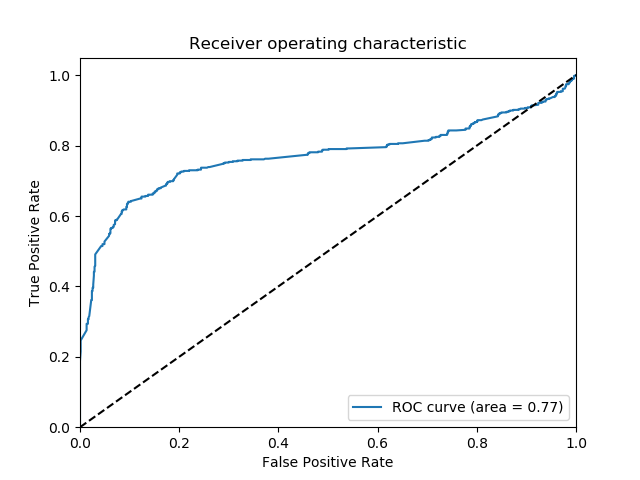}
        \caption{ROC}
        \label{fig:roc}
    \end{subfigure}
    \caption{TCN performance plots.}
    \label{fig:tcn_res}
\end{figure}

In the computational experiments, we compare performance of several models, decision tree, random forest, AdaBoost, and LSTM, along with our TCN. 
Table \ref{tab:test_metrics} illustrates the evaluation results on the test set, from which we can see that with bagging or boosting, the random forest model and AdaBoost model perform better than the decision tree model, demonstrates the power of ensemble. 
However, tree models do not capture the temporal dependencies, thus perform worse.
The deep learning models, on the other hand, significantly improve the performance to a different level, with an emphasis on learning the time-dependent patterns. 
Our TCN exhibits the highest accuracy, recall, F1 score as well as the area under the ROC curve (AUC). This demonstrates that the application of such networks can improve the prediction of sepsis in intensive care, and thus potentially further reduce sepsis mortality. In particular, it shows stronger ability to extract the temporal features than LSTM. 
The LSTM model does achieve the highest precision. However, consider the high mortality rate of sepsis, a higher recall might be more valuable for a predictive model on sepsis, as it would be safer to have less false negatives. 
We show the confusion matrix and the ROC curve obtained by our TCN model in Figure \ref{fig:tcn_res}, from which we can observe that our TCN model has a recall rate of 62\%. And as shown in the ROC curve, we may accept to sacrifice a little bit on a very high accuracy to achieve lower false negative rate, as the data are expected to be extremely imbalanced.

\begin{table}[bth!]
    \centering 
    \footnotesize
    \begin{tabular}{|c||c|c|c|c|c|c|}    \hline
 & Accuracy & Precision  & Recall & F1 Score & AUC & MCC \\ \hline \hline
Decision Tree & 0.667 & 0.668 & 0.421 & 0.517 & 0.634	& 0.300 \\ \hline 
Random Forest & 0.677 & 0.669 & 0.458 & 0.544 & 0.646 & 0.319 \\ \hline 
AdaBoost & 0.695 & 0.681 & 0.522 & 0.591 & 0.672 & 0.362 \\ \hline 
LSTM & 0.779 & \textbf{0.857} & 0.571 & 0.685 & 0.754 & 0.551 \\ \hline
TCN & \textbf{0.785} & 0.829 & \textbf{0.619} & \textbf{0.700} &  \textbf{0.770} & \textbf{0.558}
\\ \hline
    \end{tabular}
    \caption{\label{tab:test_metrics} Evaluation metrics comparison of different models on test data.}
\end{table}

Although we would not provide a comparative analysis here, during the hyperparameter tuning process, we did find that increasing the number of convolutional layers (which means the receptive fields are enlarged) in TCN would improve the model performance. This might explain the reason that TCN outperforms LSTM, as it is well-known that LSTM can hardly capture information maybe 40 time steps ago (source). And it also indicates that an improved prediction based on a longer look back might strengthen the knowledge about sepsis. As argued in Scherpf et al.\cite{scherpf2019predicting}, the symptoms and related vital sign patterns of sepsis appear quite early, and effective deep learning algorithms such as our TCN are capable of detecting such complex interdependencies between different physiological parameters.



\section{Conclusion and Discussion} \label{sec:conc}
In this paper, we build a temporal convolutional neural network to achieve the binary classification task for sepsis prediction. 
The experimental results not only show that deep learning model improves the forecasting performance significantly over several traditional machine learning models such as random forest and AdaBoost, also emphasize the value of temporal information and a gradual development of sepsis. In particular, we demonstrate that the convolutional layers have several advantages over the most widely used recurrent neural network layers  for temporal data. 
There are several directions that we can dive deeper as the future work. First of all, our dataset is relatively small for deep learning models, moreover, we only extracted 9 features from the MIMIC-III data. With a better understanding of the domain knowledge, larger dataset is expected to be obtained and more relevant features are expected to be derived, so that more useful information can be provided to increase the predictive ability of our models.
In addition, the layers/modules to learn time-dependent patterns in our deep learning models could be replaced by any other type, for instance, the recent advances of attention models \cite{vaswani2017attention} could be a good candidate. And finally, as the labels for sepsis are usually extremely imbalanced, techniques to resolve the issue of imbalanced data could be applied, for instance, the use of focal loss \cite{lin2017focal} might improve our model performance. 

\makeatletter
\renewcommand{\@biblabel}[1]{\hfill #1.}
\makeatother

\bibliography{ref}
\bibliographystyle{unsrt}


\end{document}